\documentclass[letterpaper,twocolumn,10pt]{article}
\usepackage{usenix-2020-09}
\usepackage[square,numbers]{natbib}
\usepackage{tikz}
\usepackage{amsmath}

\usepackage{multirow} %
\usepackage{bm}
\usepackage{pifont}
\usepackage{algorithm}%
\usepackage{algcompatible}%

\usepackage{subfigure}
\usepackage{array}
\usepackage{longtable}
\usepackage{bigstrut}
\usepackage{makecell}
\usepackage{enumitem}
\usepackage{booktabs}
\usepackage{threeparttable}

\usepackage{xcolor,colortbl}
\usepackage{xurl,soul}
\usepackage{hhline}
\usepackage{gensymb}

\usepackage{listings}
\definecolor{codegreen}{rgb}{0,0.6,0}
\definecolor{codegray}{rgb}{0.5,0.5,0.5}
\definecolor{codepurple}{rgb}{0.58,0,0.82}
\definecolor{backcolour}{rgb}{0.95,0.95,0.92}
\usepackage{newfloat}
\usepackage{caption}

\DeclareFloatingEnvironment[fileext=loc,placement={!ht},within=none,name=Code]{Code}

\lstdefinestyle{mystyle}{
	backgroundcolor=\color{backcolour},
	commentstyle=\color{codegreen},
	keywordstyle=\color{magenta},
	numberstyle=\tiny\color{codegray},
	stringstyle=\color{codepurple},
	basicstyle=\ttfamily\footnotesize,
	breakatwhitespace=false,
	breaklines=true,
	captionpos=b,
	keepspaces=true,
	numbers=left,
	numbersep=5pt,
	showspaces=false,
	showstringspaces=false,
	showtabs=false,
	tabsize=2
}
\begin{document}

\title{AsyncFlow: An Asynchronous Streaming RL Framework for Efficient LLM Post-Training}

\author{\rm{Zhenyu Han$\ast$ $\dagger$, Ansheng You$\ast$ $\ddagger$, Haibo Wang$\dagger$, Kui Luo$\dagger$, Guang Yang$\dagger$,}\\
\rm{Wenqi Shi$\ddagger$, Menglong Chen$\dagger$, Sicheng Zhang$\dagger$, Zeshun Lan$\dagger$, Chunshi Deng$\dagger$,}\\
\rm{Huazhong Ji$\dagger$, Wenjie Liu$\dagger$, Yu Huang$\dagger$, Yixiang Zhang$\dagger$, Chenyi Pan$\dagger$,}\\
\rm{Jing Wang$\dagger$, Xin Huang$\dagger$, Chunsheng Li$\dagger$, Jianping Wu$\dagger$ \S}\\
$\dagger$ Huawei \qquad $\ddagger$ Individual Researcher\\
\texttt{wujianping5@huawei.com}
}
\date{}
\maketitle

\begin{abstract}
Reinforcement learning (RL) has become a pivotal technology in the post-training phase of large language models (LLMs). Traditional task-collocated RL frameworks suffer from significant scalability bottlenecks, while task-separated RL frameworks face challenges in managing complex dataflows and resolving resource idling. Furthermore, most existing frameworks are tightly coupled with LLM training or inference engines, making them difficult to support custom-designed engines. To address these challenges, we propose AsyncFlow, an asynchronous streaming RL framework tailored for efficient post-training. Specifically, we introduce a distributed data storage and transfer module that provides panoramic data management and fine-grained scheduling capabilities in a fully streamed manner. This architecture inherently enables automated pipeline overlapping among RL tasks and dynamic load-balancing. Moreover, we propose an asynchronous producer-consumer workflow, which is engineered to minimize computational idleness by strategically deferring the parameter update process within staleness thresholds. Finally, the core capabilities of AsyncFlow are architecturally decoupled from underlying training and inference engines and encapsulated by service-oriented user interfaces, offering a modular and customizable user experience. Extensive experiments demonstrate an average throughput of 1.59$\times$ compared to the state-of-the-art baseline. The architecture presented in this work provides actionable insights for designing next-generation RL training systems.

\end{abstract}

%!TEX root = main.tex
\newcommand\blfootnote[1]{%
	\begingroup
	\renewcommand\thefootnote{}\footnote{#1}%
	\addtocounter{footnote}{-1}%
	\endgroup
}

\blfootnote{$\ast$ These authors contribute to this research equally.}
\blfootnote{\S ~~Corresponding authors.}
\section{Introduction}\label{sec:Introduction}
Large language models (LLMs) have demonstrated remarkable linguistic capabilities through unsupervised next-token prediction trained on massive natural language corpora \cite{radford2018improving}, paving the way for the pursuit of artificial general intelligence (AGI). Fueled by readily accessible web-scale natural language resources, model parameters have grown exponentially from millions \cite{vaswani2017attention,devlin2019bert} to trillions \cite{liu2024deepseek,grattafiori2024llama}, accompanied by continued improvement of intelligence. However, as the cornerstone of large language model development, these scaling laws \cite{kaplan2020scaling} have encountered critical data constraints as pre-training corpora are approaching exhaustion \cite{villalobos2024position,muennighoff2023scaling}. Estimates indicate that the effective stock of publicly available text data will be depleted by 2028 \cite{villalobos2024position}, threatening to halt the progress of pre-training scaling.

To enhance LLM capabilities, instruction tuning \cite{wei2021finetuned} and reinforcement learning from human feedback (RLHF) \cite{ouyang2022training} are introduced to align models with human preferences and societal values, termed as post-training process \cite{zhao2023survey}. Beyond alignment, reinforcement learning also offers a promising way to overcome data scarcity by leveraging self-generated high-quality responses and reward signals as data flywheels \cite{nvdataflywheel}, enabling iterative performance improvements. The emergence of reasoning models \cite{openaio1,team2025kimi,guo2025deepseek}, which exhibit human-level reasoning capabilities through RL-driven optimization, unveiling the post-training scaling laws \cite{sascalinglaws}. Unlike the pre-training process that only incorporates a single LLM, RL workflows involve several models and tasks that pose significant challenges for training system design. For instance, the widely used Proximal Policy Optimization (PPO) algorithm \cite{schulman2017proximal} in LLM post-training incorporates six distinct tasks: \emph{actor rollout, reference inference, critic inference, reward inference, actor update}, and \emph{critic update}. Such cross-model collaboration complicates the design of efficient and scalable RL post-training systems.

Existing RL post-training frameworks can be categorized as \textbf{task-collocated} or \textbf{task-separated}, depending on the task placement strategies. A typical task-collocated framework (e.g., DeepSpeed-Chat \cite{yao2023deepspeed}) assigns all tasks to the same set of devices. During training, only one task runs at a time, occupying all computational resources until completion. This paradigm suffers from several critical drawbacks. \emph{Memory inefficiency}. Storing all model parameters in GPU memory introduces excessive memory overhead, either limiting training efficiency or incurring costly memory offloading overhead. \emph{Resharding overhead}. Actor rollout and actor update require distinct parallelism strategies. Transitioning between these tasks necessitates model resharding, which introduces significant latency. \emph{Suboptimal Parallelism}. When handling models of varying sizes, smaller models may fail to fully utilize all computational resources due to scaling inefficiencies.

\begin{figure}[t]
        \centering
	\includegraphics[width=\linewidth]{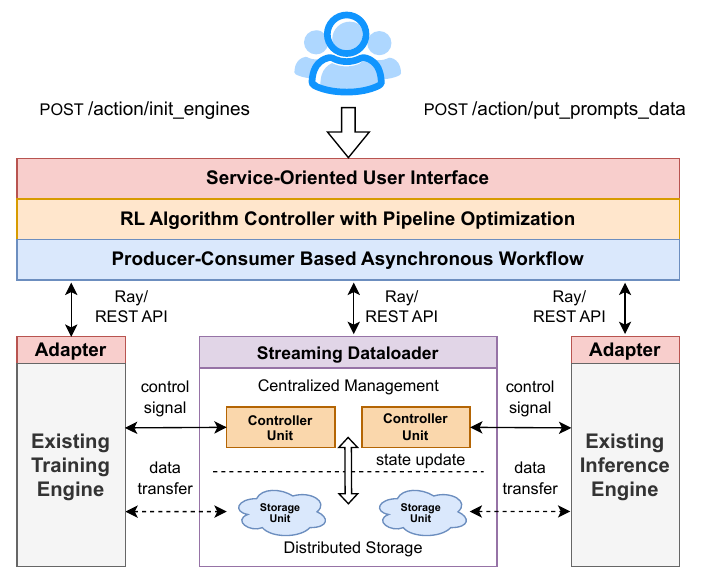}
	\caption{System overview of AsyncFlow framework.}
	\label{fig:illustration}
\end{figure}

While for task-separated frameworks, distinct RL tasks are assigned to different resource pools proportionally sized to their workload demands. In typical task-separated frameworks (e.g., OpenRLHF \cite{hu2024openrlhf}), distributed resource management and execution tools (e.g., Ray \cite{moritz2018ray}) are introduced to schedule RL workflows. However, inherent data dependencies in RL algorithms restrict full parallelization: tasks such as reference inference and critic update must wait for actor rollout completion, causing prolonged resource idling that delays the post-training process. Compounding this issue, the nested parallelism strategies of LLMs complicate cross-task dataflow between minimal instances, further degrading computational efficiency and bringing extra difficulty in algorithm development. Despite these challenges, task-separated frameworks are rapidly emerging as the community's preferred paradigm for scalable RL training, evidenced by their adoption in emerging frameworks \cite{zeng2025glm,wang2025distflow,fu2025areal,wang2025seamlessflow}.

Another critical obstacle lies in framework usability. Most existing RL frameworks are tightly coupled with specific training and inference engines. For example, NeMo-Aligner \cite{shen2024nemo} can only operate on Megatron-LM \cite{shoeybi2019megatron} and TensorRT-LLM \cite{nvtensorrt}, while OpenRLHF requires DeepSpeed \cite{rasley2020deepspeed} and vLLM \cite{kwon2023efficient} as mandatory backends. This design paradigm significantly restricts flexibility and adaptability, particularly for industrial users who often rely on pre-existing training and inference clusters built on customized backends. Ideally, an RL framework should provide high-level abstractions that seamlessly integrate with heterogeneous training and inference engines, thereby accommodating diverse customization demands. Furthermore, the absence of a unified algorithm controller exacerbates developing overhead and impedes academic research workflows that require frequent algorithm modifications.

To address these challenges, we propose AsyncFlow, an asynchronous RL framework with task-separated architecture, built atop the Ascend-powered post-training framework MindSpeed-RL \cite{msrl}. Specifically, we design a distributed data storage and transfer module named TransferQueue that optimizes dataflow across instances (\S\ref{sec:TransferQueue}). It collects all the responses produced by the inference engine, and dynamically dispatches these samples to instances of downstream RL tasks upon requests, enabling a streamed dataflow control. It eliminates the need to explicitly define cross-instance data dependency chains, thereby enabling automated load-balancing and pipeline overlapping among RL tasks. Furthermore, to fully unlock the architectural potential of task-separated framework, we propose a producer-consumer-based asynchronous workflow optimization mechanism (\S\ref{sec:pipeline_opt}). By deferring the parameter update process for rollout instances, we achieve a stable asynchronous workflow that minimizes the idle time with controllable staleness. Combined with dynamic load-balancing strategies, the efficiency of post-training can be significantly enhanced. Moreover, AsyncFlow serves as a high-level scheduling layer of RL algorithms, encapsulated by minimal service-oriented interfaces to suit various backend engines. This design principle enables an efficient and customizable user experience, which would reconcile research flexibility with industrial deployment requirements, bridging the gap between theoretical exploration and industrial applications (\S\ref{sec:user_interface}). Our core component TransferQueue is publicly available at \url{https://github.com/Ascend/TransferQueue}. %

Our contributions are summarized as follows:

\begin{itemize}
	\item We design a distributed data management module that enables automated load-balancing and pipeline overlapping among RL tasks, which greatly improves the training efficiency of task-separated RL frameworks.
	\item We propose a producer-consumer-based asynchronous workflow that can well balance the training efficiency and the convergence of the RL algorithm.
	\item We present a user-friendly service-oriented API that supports various training/inference backend engines, bridging the gap between academic research requirements and industrial deployment scalability.
	\item We conduct extensive experiments against the state-of-the-art RL post-training framework, and observe a maximum 2.03$\times$ throughput improvement over the baseline in large-scale clusters.
\end{itemize}

%!TEX root = main.tex

\section{Motivation}\label{sec:Motivation}
This section begins by introducing the foundational concepts of RL post-training for LLMs (\S\ref{sec:Motivation-background}), followed by a justification of the design choices underlying task-separated post-training frameworks (\S\ref{sec:Motivation-reason}). We conclude by identifying key challenges inherent in the task-separated paradigm (\S\ref{sec:Motivation-challenge}).

\subsection{Background}\label{sec:Motivation-background}
\subsubsection{Computational characteristic of RL post-training}
\begin{figure}[t]
        \centering
	\includegraphics[width=0.43\textwidth]{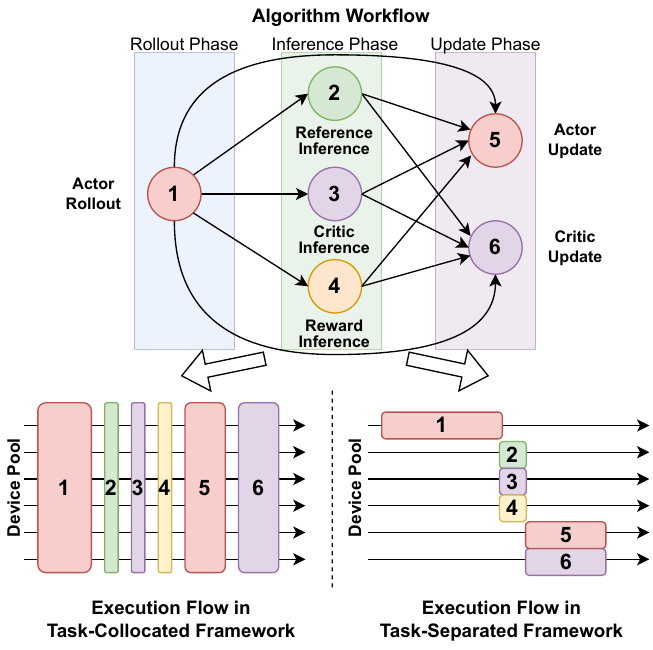}
	\caption{Algorithm and execution workflow of the PPO algorithm. Color represents different model parameters.}
	\label{fig:background}
\end{figure}

As illustrated in Figure \ref{fig:background}, RL post-training comprises three phases. The \textit{rollout phase} employs the actor model (target LLM to be trained) to generate responses to the given prompts through an auto-regressive manner, thereby producing training corpora. Subsequently, the \textit{inference phase} utilizes reward and other models to evaluate the generated responses, generating several signals that guide the RL algorithm. Finally, the \textit{update phase} optimizes the model weights based on the collected data, after which the entire training cycle repeats. Notably, the decoding process in rollout phase is \textbf{memory-bound} due to its auto-regressive nature \cite{kwon2023efficient}, while the inference phase and update phase are \textbf{computation-bound}.

Within these three phases, several models cooperate with each other, leading to complex data dependencies. In the PPO algorithm, four distinct models interact with intricate data dependencies: responses generated during actor rollout serve as input to all other models, while outputs from each model during inference phase are required by all the models within update phase. More importantly, different reinforcement learning algorithms impose unique dependencies. When combined with heterogeneous parallelism strategies (e.g., tensor parallelism \cite{shoeybi2019megatron} and pipeline parallelism \cite{huang2019gpipe}), these characteristics lead to highly complex data flow patterns in post-training systems.

\subsubsection{Architecture taxonomy of RL frameworks}
To execute reinforcement learning algorithms on physical devices, existing frameworks adopt two distinct architectures: \textbf{task-collocated} and \textbf{task-separated}. In task-collocated frameworks, all RL tasks execute sequentially across the entire cluster, forming a time division multiplexing system where hardware resources are shared through temporal scheduling. While in task-separated frameworks, RL tasks will be deployed across distinct devices—either partially or fully—establishing a spatial division multiplexing system that leverages spatial resource allocation.

\subsection{Why we need task-separated frameworks} \label{sec:Motivation-reason}
Compared to task-collocated frameworks, task-separated design offers significant advantages in both \textbf{scaling efficiency} and \textbf{deployment costs}.

First, the computational resources required for post-training are continuously increasing. For instance, the post-training phase of Grok 4 consumed as many computational resources as its pre-training, reaching a scale of 200 thousand GPUs \cite{grok4}. In this context, the disparity in computational characteristics between different tasks becomes a significant issue that hinders efficient scaling. In task-collocated frameworks, memory-bound actor rollout and compute-bound actor update phases must share the same set of devices, resulting in highly imbalanced execution time across these RL tasks \cite{zhong2025streamrl}.

Second, task-collocated frameworks incur substantial non-overlapping overhead. Specifically, transitions between RL tasks introduce significant model offloading and onloading costs, wherein the current model parameters residing on the device must be swapped out to allocate space for the next model. Even for successive tasks executed by the same model, the transition between actor rollout and actor update necessitates a complex resharding process to accommodate their divergent optimal parallelization strategies. As model size grows, these overheads become increasingly dominant, without a straightforward approach to overlap them with other computational tasks.

Third, from a cost perspective, task-separated architectures allow the RL framework to act as a top-level orchestration layer, enabling the reuse of existing training and inference clusters. Furthermore, this architecture creates opportunities to leverage heterogeneous hardware that is tailored to the divergent computational characteristics of RL tasks.

\subsection{Challenges for task-separated frameworks}\label{sec:Motivation-challenge}
By comparing the execution flows of task-collocated and task-separated frameworks (Figure \ref{fig:background}), we observe that hardware resources are significantly underutilized in task-separated scenarios, greatly limiting their scaling efficiency. To address this issue, we decompose the problem into several specific challenges.

\textbf{Dataflow management}. Inherent data dependencies between RL tasks make it difficult to achieve full parallelism in task-separated frameworks, particularly when scaling to large clusters. However, by enabling fine-grained data management, we can achieve sample-level overlapping to minimize pipeline bubbles. The design of this data management module introduces substantially greater complexity than that in task-collocated frameworks, where a simple, centralized, global-batch-level data buffer suffices.

\textbf{Asynchronous algorithm design}. Traditional RL algorithms require strict synchronization between actor rollout and actor update processes. This synchronization precludes the actor rollout from processing new prompts from the subsequent global batch, thereby introducing significant pipeline bubbles across training iterations. While asynchronous, off-policy RL algorithms can yield substantially higher efficiency, reducing the impact on algorithmic convergence remains a critical challenge.

\textbf{User-friendly interface}. Task-separated frameworks introduce inherent complexity in API design due to the physical disaggregation of computation tasks. To reduce pipeline bubbles, the execution flow of each RL task must be carefully organized, making the traditional single-controller paradigm inadequate for governing the entire post-training process. This poses significant challenges for academic researchers who are seeking to develop novel RL algorithms without extensive modifications to the post-training framework. At the same time, industrial users require easy adaptability to customized training and inference backends already deployed on existing clusters. These requirements make it difficult to provide an intuitive, easy-to-use interface.

%!TEX root = main.tex

\section{AsyncFlow Overview}\label{sec:System Overview}
\begin{figure}[t]
        \centering
	\includegraphics[width=\linewidth]{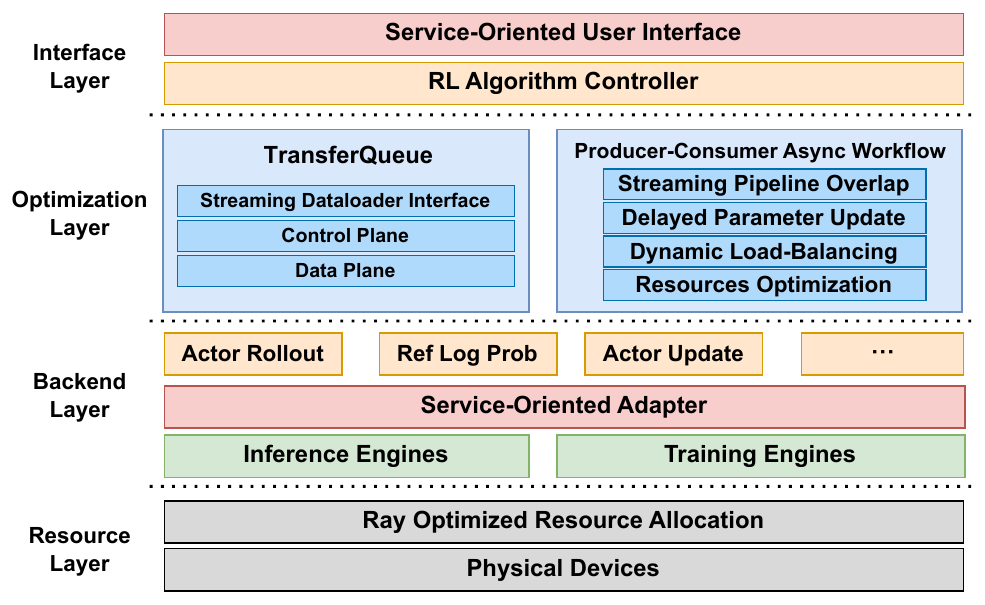}
	\caption{Hierarchical architecture design of AsyncFlow framework.}
	\label{fig:system_overview}
\end{figure}

To address the above challenges, we introduce AsyncFlow, an asynchronous streaming reinforcement learning framework for scalable post-training in Figure \ref{fig:system_overview}.

First, we present TransferQueue, an asynchronous streaming data management module that dynamically schedules complex data dependencies in post-training. Specifically, TransferQueue offers fine-grained, sample-level data management and load-balancing capabilities, upon which we implement a gateway routing mechanism that decouples explicit data dependencies across RL tasks. This enables a divide-and-conquer approach, significantly simplifying the design of the algorithm controller.

Second, we propose an asynchronous off-policy RL workflow that maximizes computational resource utilization without significantly compromising algorithmic convergence. By strategically delaying parameter updates within a staleness threshold, we allow actor rollout to process new prompts from the next global batch, reducing pipeline bubbles across iterations. Furthermore, we propose a sub-step asynchronous algorithm prototype operating in a producer-consumer pattern. By enabling a subset of actor rollout instances to perform parameter updates immediately while concurrently allowing others to serve downstream tasks, this design enhances system efficiency while mitigating convergence impact.

Third, we carefully design a user-friendly service-oriented interface that reconciles research flexibility with industrial deployment requirements. The resource layer leverages Ray for computing resource management, with hardware allocation pre-optimized through an execution time simulator to ensure efficient training. Building upon this, the backend layer provides modular adapters for diverse training and inference engines, allowing RL tasks to be instantiated on these engines while remaining engine-agnostic. For usability design, the interface layer offers a unified algorithm entry point. It serves as a single controller that satisfies the requirements of algorithm research. Complementing this, we provide a set of service-oriented APIs for industrial workflows, which enable seamless integration with existing training and inference clusters. In summary, AsyncFlow aims to bridge the gap between algorithmic research and industrial deployment in LLM post-training by delivering a unified framework that harmonizes flexibility with scalable efficiency.

%!TEX root = main.tex

\section{TransferQueue: An Asynchronous Streaming Data Management Module}\label{sec:TransferQueue}
\subsection{Architecture Overview}
The design of the data system largely determines the computational architecture. To address the system challenges identified in prior analysis, we propose TransferQueue, a high-performance data storage and transfer module featuring a panoramic view of data and streaming scheduling capabilities, to enable efficient dataflow in post-training systems.

\begin{figure}[t]
        \centering
	\includegraphics[width=\linewidth]{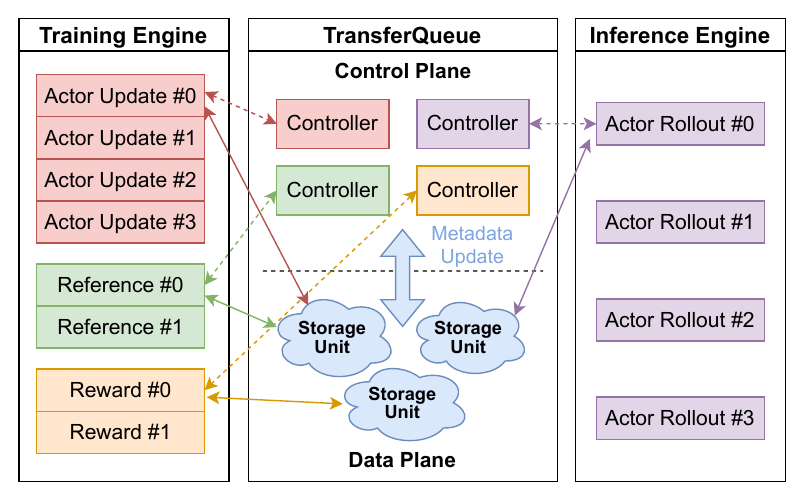}
	\caption{Architectural design of TransferQueue. Each DP group interacts with its corresponding TransferQueue controller to get the metadata upon request, then executes the read/write operation with storage units in the data plane. The dashed line represents metadata communication, while the solid line depicts the communication process with real data. All these interaction processes are encapsulated as a distributed streaming dataloader to seamlessly integrate into training and inference engines.}
	\label{fig:tq_architecture}
\end{figure}

As illustrated in Figure \ref{fig:tq_architecture}, TransferQueue bridges training and inference clusters, managing the entire dataflow within the RL post-training process. Specifically, we decouple the data plane from the control plane, instantiating multiple controllers and data storage units in each plane. This architecture mitigates potential I/O and network bottlenecks by distributing requests across parallelized units, enabling scalable post-training.

\textbf{For data management capabilities}, each TransferQueue controller copy maintains the data status for all training samples, providing a panoramic view of data status for each RL task. The data status is organized at the sample level, supporting fine-grained scheduling in a streamed manner. Functioning as a logical gateway router, TransferQueue controller dynamically redirects dataflow between RL tasks under varying load balancing policies (e.g., balancing the token counts across each DP). This design facilitates micro-batch level overlapping by enabling downstream tasks access to part of the training samples for computation, eliminating the need to wait for the entire dataset.

\textbf{For data storage capabilities}, each unit manages a subset of samples within each global batch. These samples are assigned global indices to ensure accurate addressing by all distributed controllers. When new data is written to a storage unit, it triggers an update notification to all controllers to update their data status.

\begin{figure}[t]
        \centering
	\includegraphics[width=0.42\textwidth]{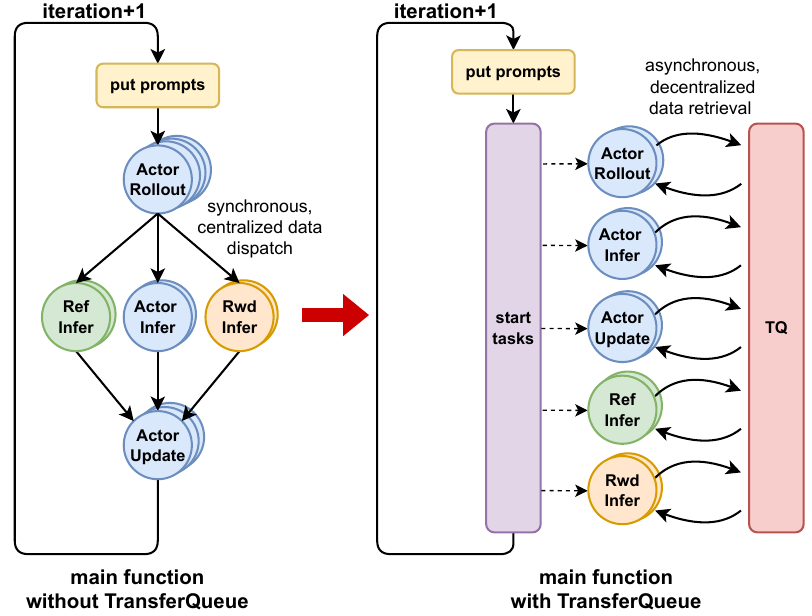}
	\caption{RL workflow without or with TransferQueue. The left panel illustrates the traditional algorithm workflow, where explicit organization of dataflow is required. With TransferQueue, RL tasks can be further decoupled, simplifying algorithm design.}
	\label{fig:controller_flow}
\end{figure}

As illustrated in Figure \ref{fig:controller_flow}, TransferQueue eliminates manual configuration of dataflows across parallel processes of multiple RL tasks. This design distinguishes our design from existing solutions such as OpenRLHF \cite{hu2024openrlhf} and StreamRL \cite{zhong2025streamrl}, offering a more flexible and efficient programming paradigm.

The TransferQueue design significantly simplifies the architectural design of RL frameworks. As illustrated in the left panel of Figure \ref{fig:controller_flow}, traditional RL workflows usually require a synchronous, centralized data dispatch mechanism within the main function. In some task-separated frameworks, we must explicitly define data dependencies across DP groups for varying RL tasks (see Appendix Code \ref{code:data_dependency_in_openrlhf}). By contrast, the right panel demonstrates how TransferQueue simplifies the design: it enables all RL tasks to be launched within the main function, with each task autonomously managing its data dependencies in a divide-and-conquer manner. This paradigm sets our framework apart from existing solutions such as OpenRLHF \cite{hu2024openrlhf} and StreamRL \cite{zhong2025streamrl}, establishing a more flexible and efficient programming model for distributed reinforcement learning systems.

\subsection{Interaction Interface} \label{sec:tq_interact}
The interaction procedure with TransferQueue can be divided into two phases: metadata retrieval and data retrieval. Metadata retrieval is handled by the algorithm controller, enabling straightforward debugging capabilities. Data retrieval is executed by distributed workers, accelerating the data retrieval.

Specifically, when an RL task requires new data, its algorithm controller (i.e., the main function) initiates the process by sending a read request to the corresponding TransferQueue controller. TransferQueue controller dynamically assembles a batch of samples from currently available data and returns their metadata to the algorithm controller. Worker processes (each bound to a single Ascend NPU) then retrieve the corresponding data from distributed storage units using the provided metadata, eliminating the centralized bottleneck that would occur if raw data were routed through the algorithm controller. To ensure each DP group accesses distinct, non-duplicate samples, TransferQueue controller tracks consumption records for each sample, guaranteeing that only one DP group within the RL task accesses the sample's metadata. Similarly, once the RL task completes its computation, it writes results back to the storage units according to the metadata. The storage units then notify all the controllers to update the data status. This mechanism ensures controllers are immediately aware that the data at these locations is ready for consumption.

To simplify the usage of TransferQueue, we have encapsulated its capabilities into a PyTorch DataLoader. As demonstrated in Appendix Code \ref{code:tq_example}, users first initialize TransferQueue as a streaming dataloader by specifying the current RL task, required input data columns, and micro-batch size. During each forward step, the required data can then be easily retrieved via an iterator wrapper. This interface simplifies TransferQueue's integration, ensuring seamless compatibility with existing training workflows.

\subsection{Data Plane: Distributed Data Storage}

\begin{figure}[t]
        \centering
	\includegraphics[width=\linewidth]{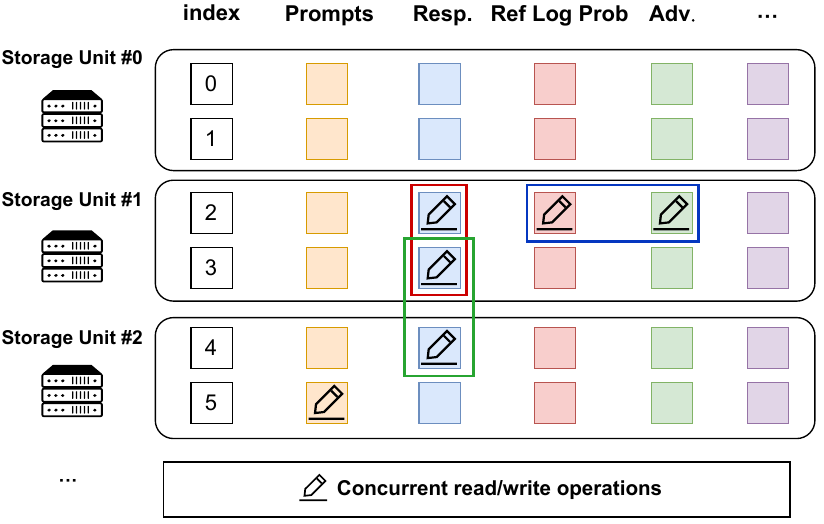}
	\caption{Data structure of TransferQueue. Each row represents a complete data sample, while each column represents the task-specific data component. Through the global index, we can accurately address the data samples across different storage units for concurrent read/write operations.}
	\label{fig:tq_data_structure}
\end{figure}

To address the diverse data requirements of RL tasks, we adopt a 2D columnar data structure, which is illustrated in Figure \ref{fig:tq_data_structure}. In this design, columns correspond to task-specific data components, such as actor responses and reference log probabilities. Rows represent complete training samples, each of which can be uniquely addressed by a global index. This structure enables RL tasks to retrieve only the specific data components according to the algorithm requirements. For instance, the reference model needing input token IDs and responses can exclusively request the relevant columns, minimizing unnecessary data transfer. Additionally, this organization supports concurrent read/write operations across distinct positions, enhancing scalability in high-concurrency scenarios.

Considering the potential I/O and bandwidth bottlenecks, we store the training samples in a distributed manner. Each of the storage units maintains a subset of data entries (i.e., rows) to balance storage and communication overhead across the system.

\subsection{Control Plane: Panoramic Data Admin}
\begin{figure}[t]
        \centering
	\includegraphics[width=0.45\textwidth]{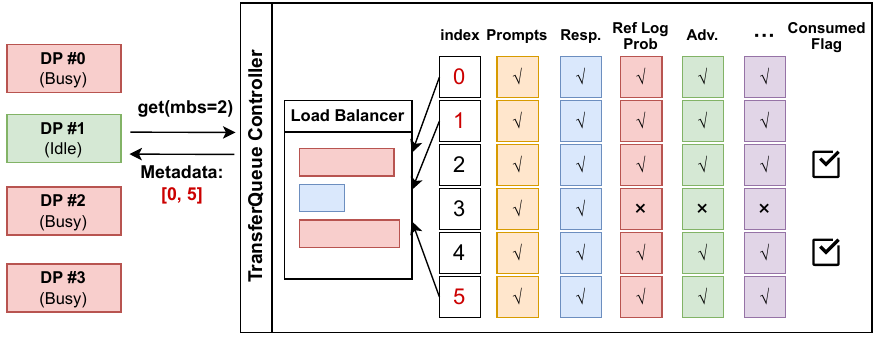}
	\caption{Procedure for a \textit{get} request in the TransferQueue controller. Red indices indicate that the corresponding data samples meet the requirements of the RL task; these samples will be sent to the load balancer, and the metadata of the selected data will be returned to the requester.}
	\label{fig:tq_control}
\end{figure}

The nested parallelisms of LLM and concurrent RL tasks create intricate dataflows during post-training. To disentangle these complex data dependencies, the TransferQueue controller acts as a gateway router with a panoramic view of data, ensuring coherent coordination across distributed workers for each task.

The TransferQueue controller maintains the data status of each training sample using binary indicators: \ding{55} indicates data unavailability, while \ding{52} denotes that the data is ready for retrieval. As described in \S\ref{sec:tq_interact}, the data status is dynamically updated when new data is written to data storage units.

As illustrated in Figure \ref{fig:tq_control}, upon receiving a read request, the TransferQueue controller scans the data status to identify entries that meet the task requirements. Specifically, the controller generates a mask containing the requested data columns, then finds entries marked entirely with \ding{52} and that have no prior consumption records from other DP groups. Then the controller runs a load-balancing algorithm to select the required number of data samples. Finally, the metadata of these selected samples is sent to the requester. Using the metadata, worker processes communicate with distributed data storage units to perform the actual data retrieval. These samples are marked as consumed to prevent duplication by other DP groups within the same RL task.

A key advantage of our architecture lies in its centralized data management. By dynamically scheduling all the data through the controller instead of pre-allocating to DP groups, TransferQueue enables advanced load-balancing strategies. Faster instances (due to hardware heterogeneity or shorter response lengths) can dynamically request more data, enhancing overall system efficiency. Furthermore, TransferQueue enables proactive load-balancing to ensure equitable distribution of processed tokens across DP and PP groups, reducing computational idling during training. Specifically, we formalize load-balancing as a one-dimensional bin packing problem. For DP load-balancing, we enforce a consistent number of samples across DP groups, while minimizing the standard deviation of total token counts. For PP load-balancing, we further relax the constraint of sample counts. The TransferQueue controller solves this problem using the near-optimal Karmarkar–Karp algorithm \cite{karmarkar1982efficient}, ensuring equitable token distribution across DP and PP groups. In summary, TransferQueue offers valuable insights for the development of next-generation dataflow management systems.

%!TEX root = main.tex

\section{Producer-Consumer-Based Asynchronous Workflow Optimization} \label{sec:pipeline_opt}
A critical challenge in task-separated frameworks stems from pipeline bubbles—idle hardware resources caused by data dependencies for tasks that are deployed across different sets of devices. To overcome this challenge, AsyncFlow implements a series of pipeline optimization techniques to maximize hardware utilization.

First, our streaming dataloader enables pipeline overlapping among RL tasks, which greatly reduces the end-to-end execution time of post-training. Second, in off-policy scenarios, we design a delayed parameter update mechanism that well balances algorithmic convergence and training efficiency. By allowing a continuous one-step asynchronization between actor rollout and actor update, this approach effectively eliminates the bubbles caused by warm-up and cool-down phases (as demonstrated in Figure \ref{fig:async_algo}). Finally, fine-grained execution time optimization is achieved through proactive resource planning, fully unlocking the scalability of task-separated frameworks in LLM post-training.

\subsection{Sample-Level Overlapping across Tasks}
\begin{figure}[t]
        \centering
	\includegraphics[width=\linewidth]{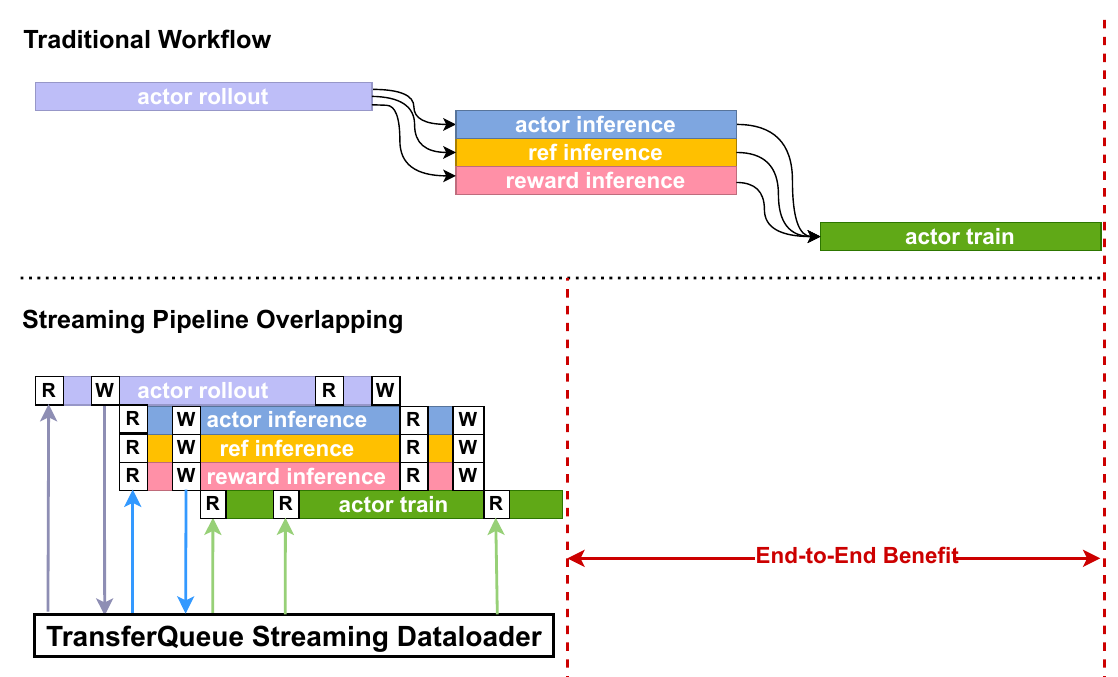}
	\caption{Streaming pipeline overlapping.}
	\label{fig:pipeline_overlapping}
\end{figure}

Leveraging TransferQueue, AsyncFlow not only simplifies dataflow management but also enhances training throughput by enabling pipeline overlapping across RL tasks. In this paradigm, all the training and inference instances only need to interact with the streaming dataloader, which dynamically schedules and redirects the finest-grained data samples across tasks (demonstrated in Figure \ref{fig:controller_flow}). Therefore, we inherently support fine-grained pipeline overlapping in Figure \ref{fig:pipeline_overlapping}. Unlike the design in existing frameworks \cite{zhong2025streamrl,hu2024openrlhf}, we can easily extend this overlapping strategy to any RL algorithms with varying tasks, eliminating the need to manually reschedule the data streams.

\subsection{Asynchronous Off-Policy Optimization}

\begin{figure*}[htbp]
  \makebox[\textwidth][c]{
    \includegraphics[width=1\textwidth]{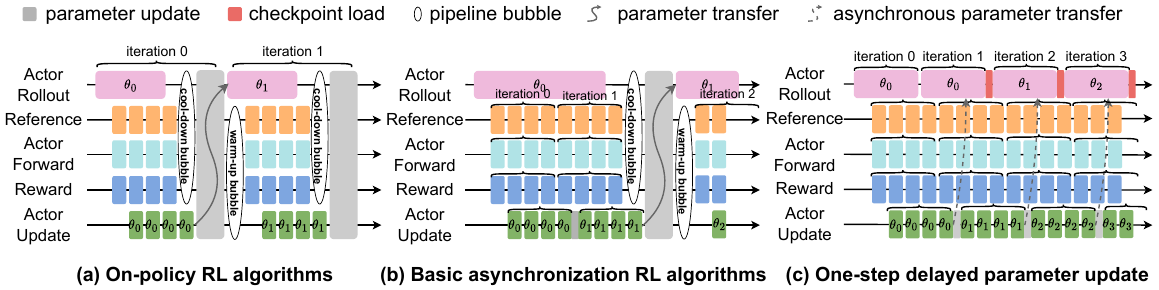}
  }
    \caption{Illustration of the asynchronous off-policy RL workflow. $\theta_\star$ represents the model version, which is trained by the data in iteration $\star$. Leveraging the proposed delay parameter update mechanism in (c), AsyncFlow can extend the steady phase of the RL pipeline to further reduce pipeline bubbles.}
    \label{fig:async_algo}
\end{figure*}

\subsubsection{Basic asynchronous RL algorithm}
Traditional RL algorithms for LLM post-training predominantly adopt a synchronous on-policy paradigm, where actor rollout and actor update operate on identical parameter states. While this ensures algorithmic convergence, its computational inefficiency—stemming from strict synchronization—severely limits scalability for large-scale post-training. Recent advancements have challenged this paradigm: optimization techniques such as partial rollout in Kimi K1.5 \cite{team2025kimi} and streaming rollout in Seed1.5-Thinking \cite{seed2025seed} practically relax the on-policy assumption by allowing response data generated by older models used in the update stage. These innovations create opportunities to explore asynchronous, off-policy frameworks, which promise significant gains in training efficiency while maintaining convergence guarantees.

As illustrated in Figure \ref{fig:async_algo}(a), on-policy algorithms enforce strict synchronization between actor rollout and actor update stages, which leads to warm-up and cool-down bubbles across iterations. A straightforward mitigation strategy involves enlarging the global batch size of actor rollout stage, transitioning to an asynchronous off-policy scenario where actor update utilizes responses generated by stale parameters, as illustrated in Figure \ref{fig:async_algo}(b). This adaptation extends the stable phase of the RL pipeline, thereby reducing the proportion of pipeline bubbles in the warm-up and cool-down phases. However, algorithmic convergence constraints impose strict limits on the allowable version differences between actor rollout and actor update stages. Empirical studies demonstrate that one-step asynchronization between actor rollout and actor update will not lead to significant performance degradation or convergence issues \cite{noukhovitch2024asynchronous, zhong2025streamrl}. However, the performance drops logarithmically with the increasing version differences beyond this threshold. This trade-off highlights the critical need for mechanisms that reconcile training efficiency with algorithmic stability in large-scale LLM post-training.

\subsubsection{Delayed parameter update mechanism}
To solve the above dilemma, we present a delayed parameter update mechanism that further eliminates pipeline bubbles.
As illustrated in Figure \ref{fig:async_algo}(c), this mechanism defers parameter update by one step, enabling continuous rollout during the transition period across iterations. Specifically, the actor rollout worker does not immediately halt generation when the actor update is completed. Instead, it continues generating responses using the old model weights while asynchronously writing the received new parameters to the host memory. The new parameters will load to the Ascend Neural Network Processing Units (NPUs) when the current generation iteration is completed, reducing the exposed synchronization overhead to a relatively fast host-to-device (H2D) transmission. By allowing a continuous one-step asynchronization, it enables the stable phase to be nearly infinitely extendable by eliminating warm-up and cool-down phases. While conceptually similar to StreamRL \cite{zhong2025streamrl}, our approach further enhances scalability by overlapping all RL tasks within the training engine through the dataflow management of TransferQueue. In summary, this design effectively reduces the warm-up and cool-down pipeline bubbles in RL post-training workflows, fully exploiting the architectural potential of task-separated frameworks.

Furthermore, we propose a novel parameter updating mechanism that enables a sub-step asynchronous algorithm workflow, as illustrated in Appendix Figure \ref{fig:async_algo_future}. To efficiently supply data for downstream tasks, we generally allocate abundant hardware resources to the actor rollout task. This setup allows rollout instances to perform parameter updates sequentially, while the remaining instances continue fulfilling the data requirements of downstream training tasks. Consequently, in each iteration, we can leverage the most recently updated parameters to generate part of the data, achieving sub-step asynchrony. In addition, this design also minimizes the overhead of checkpoint loading, thereby enhancing system efficiency. We leave the implementation of this mechanism for future work.

\subsubsection{Parameter update overlapping}
The parameter update module is mainly composed of a $WeightSender$ deployed on the training cluster and a $WeightReceiver$ on the inference cluster. To minimize the weight transmission time in the RL training pipeline, the sender and receiver are designed to support both synchronous and asynchronous modes.

In synchronous mode, actor rollout is blocked by the parameter update process. To shorten the exposed transmission time, we leverage high-bandwidth Huawei Collective Communication Library (HCCL) links across Ascend NPUs to transmit the model weights. We also develop an asynchronous parameter update process that can fully overlap with computation tasks for asynchronous RL algorithms. Specifically, model weights from the training engine are offloaded to the host device and asynchronously transmitted to the inference engine over the host network, decoupling computational workloads from the weight synchronization process. This design ensures that the parameter update process neither stalls nor interferes with ongoing computational tasks, maintaining a continuous RL workflow.

\subsection{Task Resource Planning}
To achieve the ideal workflow illustrated in Figure \ref{fig:async_algo}, we build a graph-based resource planning module that accurately searches the optimal configuration under given resource constraints. Through simulating the computational and communication time under varying configurations, we can acquire the resource allocation setting and hyper-parameters that minimize the end-to-end time of the whole RL workflow.

Given the large search space for LLM post-training, we adopt a hybrid cost model combining an analytical-based method and a profiling-based method. The analytical-based method estimates the execution time leveraging the hardware specifications and theoretical computation and communication volumes, offering a fast evaluation that can quickly narrow down the search space. In contrast, the profiling-based method provides block-level performance by actually running training and inference tasks. It provides an accurate evaluation at a much higher time consumption. Leveraging the hybrid cost model, we ensure a good balance between efficiency and accuracy, making it well-suited for optimizing the complex RL workflows for LLM post-training.

%!TEX root = main.tex

\section{Service-Oriented User Interface} \label{sec:user_interface}
To deliver an enhanced user experience, RL frameworks should provide a high-level abstraction layer that orchestrates all the algorithm-specific tasks. For academic research, this design offers a unified entry point for algorithm development, enabling rapid experimentation through standardized APIs. For industrial scenarios, it is critical to support various training and inference backends while maintaining architectural flexibility. Therefore, maintaining a proper user interface is of critical importance.

\begin{figure}[t]
        \centering
	\includegraphics[width=\linewidth]{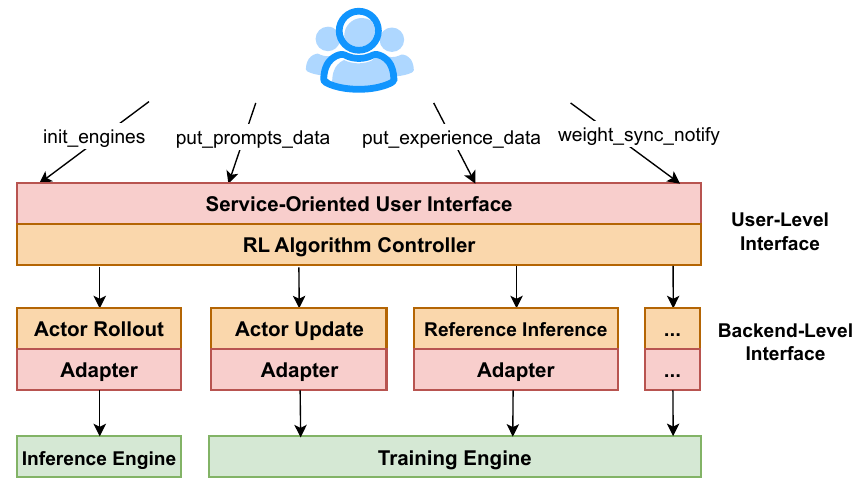}
	\caption{Architecture of service-oriented user interface.}
	\label{fig:service_api}
\end{figure}

In AsyncFlow, we implement a hierarchical service-oriented interface shown in Figure \ref{fig:service_api}. Specifically, the user-level interface encapsulates RL algorithm logics for end-users, exposing a set of key APIs to control the post-training workflow. Besides, the backend-level interface offers a modular abstraction of RL tasks, decoupling the algorithm logic from execution engines through several backend adapters. The above design achieves a clear separation of concerns: algorithm researchers can interact with high-level APIs for algorithm designs, while infrastructure engineers can focus on low-level implementation to achieve higher efficiency. The synergy of these interfaces balances academic flexibility with industrial scalability, ensuring that AsyncFlow serves both rapid algorithm iteration and production-grade deployment demands.

\subsection{User-Level Interface}

AsyncFlow provides an RL algorithm controller within the $Trainer$ class, serving as the centralized entry point for the main training workflow. This abstraction seamlessly organizes critical RL tasks, such as $generate\_sequences$ and $update$. Researchers can easily modify the core RL algorithms through the $Trainer$ class. To facilitate workflow automation in industrial scenarios, we implement several key APIs for initiating the full post-training task with minimal configuration, such as:

\begin{itemize}
	\item $init\_engines$: Initialize training and inference engines.
	\item $put\_prompts\_data$: Load the prompt dataset to the post-training system.
	\item $put\_experience\_data$ and  $get\_experience\_data$: Coordinate the generated experience data between training and inference engines.
	\item $weight\_sync\_notify$: Notify the training and inference engine to update the model weights.
\end{itemize}

The above user-level interface offers intuitive control over the post-training workflow, thereby streamlining system utilization.

\begin{figure*}[t]
        \centering
	\includegraphics[width=0.9\textwidth]{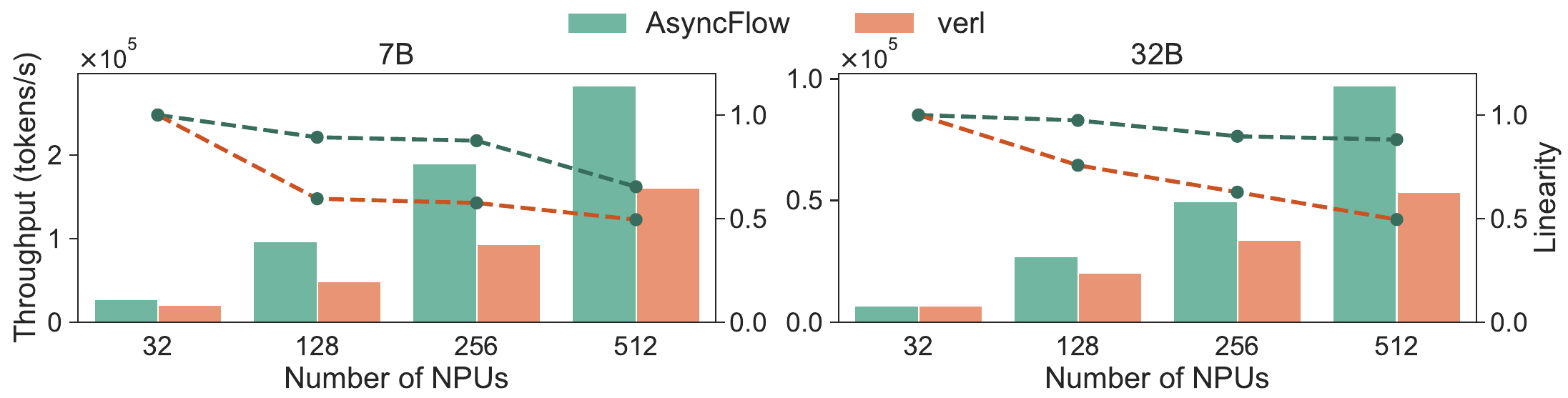}
	\caption{End-to-end throughput and scalability analysis across varying cluster and model sizes. We report the average throughput across 10-20 iterations to reduce measurement fluctuations at the beginning.}
	\label{fig:main_result}
\end{figure*}

\subsection{Backend-Level Interface}

Considering the varying training and inference backends, we design a low-level abstraction of RL tasks through the $Adapter$ class, which is demonstrated in Appendix Code \ref{code:backend_interface_example}.

This layer ensures seamless integration of diverse training/inference backends (e.g., FSDP, DeepSpeed, vLLM) while maintaining compatibility with custom-designed frameworks, allowing engineers to optimize hardware utilization or switch systems without disrupting workflow integrity.

%!TEX root = main.tex

\section{Evaluation}\label{sec:Evaluation}

\subsection{Experiment Setup}

\textbf{Models}. During the evaluation of AsyncFlow framework, we chose Qwen2.5 series of models \cite{yang2024qwen2}, ranging from 7B to 32B parameters. Qwen2.5 is widely used for both academic research and industrial workflows, serving as a common base model for RL post-training evaluation \cite{zhong2025streamrl}.

\textbf{RL algorithms}. We implement Group Relative Policy Optimization (GRPO) \cite{shao2024deepseekmath} as a representative RL algorithm for evaluation. GRPO eliminates the need for a separate critic model by estimating advantages from rewards normalized within a group of responses to the same prompt. This design greatly improves post-training efficiency, as demonstrated by its successful application in DeepSeek-R1 \cite{guo2025deepseek}. Support for PPO (Proximal Policy Optimization) \cite{schulman2017proximal} is currently under development. %

\textbf{Datasets}.
We adopt the DeepScaleR dataset \cite{deepscaler2025} for RL post-training. It contains more than 40 thousand mathematics question-answer-solution pairs that are carefully compiled from AIME (American Invitational Mathematics Examination), AMC (American Mathematics Competition), etc. Leveraging this dataset, DeepScaleR-1.5B-Preview has surpassed the performance of OpenAI o1-Preview \cite{openaio1} in several benchmarks, demonstrating the effectiveness of the dataset. In recently developed RL post-training frameworks \cite{fu2025areal}, DeepScaleR acts as a common dataset choice.

\textbf{Hardware configuration and parallelism strategies}. We use large-scale clusters with Ascend Neural Network Processing Units (NPUs) to evaluate the proposed RL framework. Each node has 16 NPUs, with system memory of 2880 GB.

\textbf{Software versions}. We use Ascend Extension for PyTorch 7.0.0 (PyTorch-2.5.1) and Ascend Compute Architecture for Neural Networks (CANN) 8.1.RC1 to serve as the basic software platform for NPU support. In AsyncFlow, we use vLLM-Ascend 0.7.3 as the inference backend, along with MindSpeed as the training backend.

\textbf{Baselines}. We evaluate the proposed AsyncFlow framework against verl \cite{sheng2024hybridflow}, the state-of-the-art RL framework that employs an efficient 3D-HybridEngine to minimize resharding overhead. Besides the high training efficiency, its combination of single-controller and multi-controller paradigms also greatly simplifies the software development. We also assessed OpenRLHF \cite{hu2024openrlhf}, a representative task-separated RL framework; however, given verl's established $3.25\times$ average performance advantage in PPO algorithm \cite{sheng2024hybridflow}, re-evaluation became less critical. Our experiments find that OpenRLHF's NVIDIA-optimized implementation exhibits suboptimal out-of-the-box performance on Ascend NPUs, leading us to exclude its results from this study. For emerging frameworks like AReaL \cite{fu2025areal} and Slime \cite{zeng2025glm}, the absence of native Ascend support and their frequent updates made fair comparison impractical within our experimental scope. During the experiment, we chose Pytorch FSDP as the training backend and vLLM-ascend 0.7.3 \cite{kwon2023efficient} as the inference backend for verl. To run on Ascend NPU platform, we made necessary adaptations to verl (version dated April 7, 2025, with commit d13434f).

\subsection{Overall Performance Analysis}
We conduct extensive experiments on clusters scaling from 32 to 512 NPUs, evaluating the performance of AsyncFlow on Qwen2.5-7B and Qwen2.5-32B models. As shown in Figure \ref{fig:main_result}, AsyncFlow consistently outperforms the verl baseline across all configurations, achieving an average throughput gain of 1.59$\times$. The performance improvement is most notable at scale: a peak speedup of 2.03$\times$ is achieved for the 7B model on 256 NPUs, while throughput remains 1.76$\times$(7B) and 1.82$\times$(32B) over verl on 512 NPUs. This phenomenon is in line with previous studies, in which task-separated frameworks show a greater potential in large-scale scenarios \cite{sheng2024hybridflow,zhong2025streamrl}
by employing parallelism strategies tailored to RL tasks. Even on 32 NPUs, AsyncFlow maintains a $33.4\%$ higher throughput for 7B model, highlighting its robust adaptability to resource-constrained environments. Such an advantage can be attributed to the streaming data scheduling and full pipeline overlapping across RL tasks and parameter updates, which effectively maximizes resource utilization.

Critically, AsyncFlow sustains a significantly higher scaling efficiency against verl, maintaining a linearity of 0.65 on 7B model and 0.88 on 32B model when cluster size expands 16$\times$.
Such strong scalability underscores a breakthrough with AsyncFlow, which paves the way for efficient training of LLM reasoning agents at industrial scales.

\subsection{Ablation Studies}

\begin{table}
	\caption{Performance improvement breakdown for 7B model running on 512 NPUs.}
	\label{tab:ablation}
	\begin{tabular}{ccccc}
		\toprule
		No. & Setting &\# Normalized Throughput \\
		\midrule
		\ding{192}& Baseline & 1 \\
		\ding{193}& w/TransferQueue & 2.01\\
		\ding{194}& \ding{193} + w/Asyn.Opt &  2.74\\
		\bottomrule
	\end{tabular}
	\vspace{-0.1in}
\end{table}

To validate the effectiveness of the proposed methods, we conduct a series of ablation studies summarized in Table \ref{tab:ablation}.

\textbf{Baseline}. We establish a baseline scenario representing a conventional task-separated RL framework by disabling all proposed optimizations. In this scenario, each RL task is allocated to dedicated hardware resources, and only one task executes at any given time. This sequential workflow is illustrated at the top of Figure \ref{fig:pipeline_overlapping}.

\textbf{TransferQueue}. As a core feature of AsyncFlow, TransferQueue enables fine-grained overlapping across RL tasks. Compared to the baseline, integrating TransferQueue yields a 2.01$\times$ throughput.

\textbf{Asynchronous workflow optimization}. To minimize the idle periods across iterations, we build an optimized asynchronous workflow. It includes the delayed parameter update mechanism, overlapping techniques, and task resource allocation strategies. The optimization further improves throughput by 36.3\% over the TransferQueue-enabled baseline.

\subsection{Load-Balancing Strategies}
In this section, we evaluate the proposed load-balancing strategies for \emph{reference inference} (forward-only) and \emph{actor update} (forward-backward) phases. As shown in Table \ref{tab:lb}, PP load-balancing achieves $2.91\times$ and $1.81\times$ throughput improvements over baseline for \emph{reference inference} and \emph{actor update} respectively. Moreover, when combined with DP load-balancing, throughput further increases to $3.03\times$ and $1.85\times$.

This performance gain reveals a fundamental challenge of RL post-training: the high variance in response lengths invalidates static sample partitioning at system initialization. Consequently, dynamic runtime scheduling becomes essential to efficiently manage the dataflow. Leveraging TransferQueue's architecture, we can dynamically allocate workloads based on real-time demands.

\begin{table}
	\caption{Normalized throughput under load-balancing strategies for 7B model running on 16 NPUs.}
	\label{tab:lb}
	\setlength{\tabcolsep}{5pt}
	\begin{tabular}{ccc}
		\toprule
		Setting & Reference & Actor Update \\
		\midrule
		Baseline& 1 & 1 \\
		w/PP Load-Balancing& 2.91 & 1.81\\
		w/PP \& DP Load-Balancing& 3.03 &  1.85\\
		\bottomrule
	\end{tabular}
	\vspace{-0.1in}
\end{table}

\subsection{Optimized Workflow of AsyncFlow}
\begin{figure}[t]
        \centering
	\includegraphics[width=\linewidth]{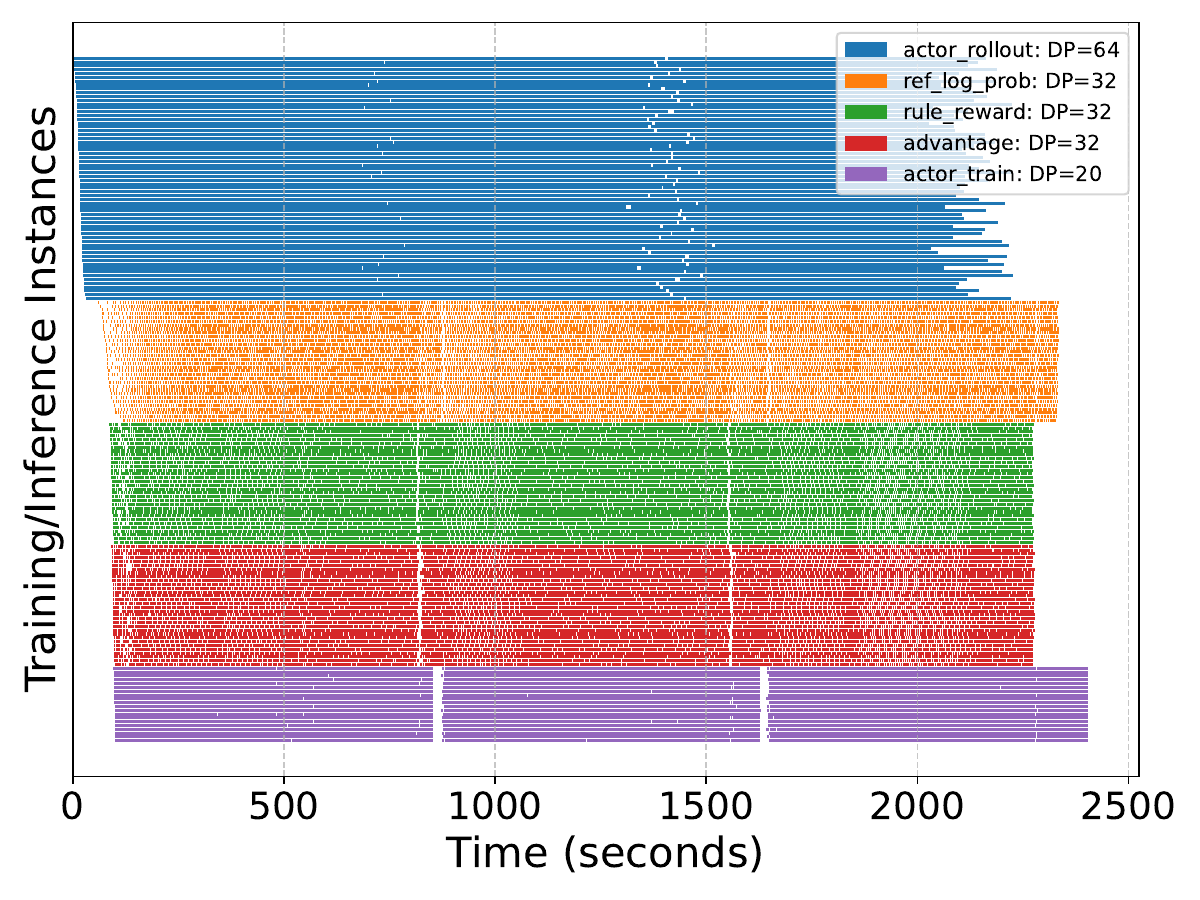}
	\caption{AsyncFlow workflow for each training and inference instance. We showcase the 32B model with 512 NPUs, with iterations 0-3. Each row on the figure represents one DP of its corresponding RL task and each block depicts the processing of one micro-batch data.}
	\label{fig:workflow_in_practice}
\end{figure}

To rigorously evaluate the efficacy of the proposed framework, we present the empirical execution timeline of distributed training and inference instances in Figure \ref{fig:workflow_in_practice}.

Specifically, the task-separated architecture enables each RL component to operate at its optimal parallelism level. For instance, the \textit{actor rollout} is allocated the highest data parallelism (DP=64) to maximize generation throughput, thereby reducing inter-task latency for downstream tasks, while the \textit{actor update} employs a lower data parallelism size (DP=20), as its computational speed aligns with the generation rate of the actor rollout. The highly overlapping and nearly contiguous execution across all tasks demonstrates that our micro-batch-level dataflow scheduling achieves substantial parallelism with minimal inter-task idle time. This outcome confirms that our task-separated RL framework effectively balances resource utilization and scalability, enabling efficient post-training in large-scale settings.

\begin{figure}[t]
        \centering
	\includegraphics[width=\linewidth]{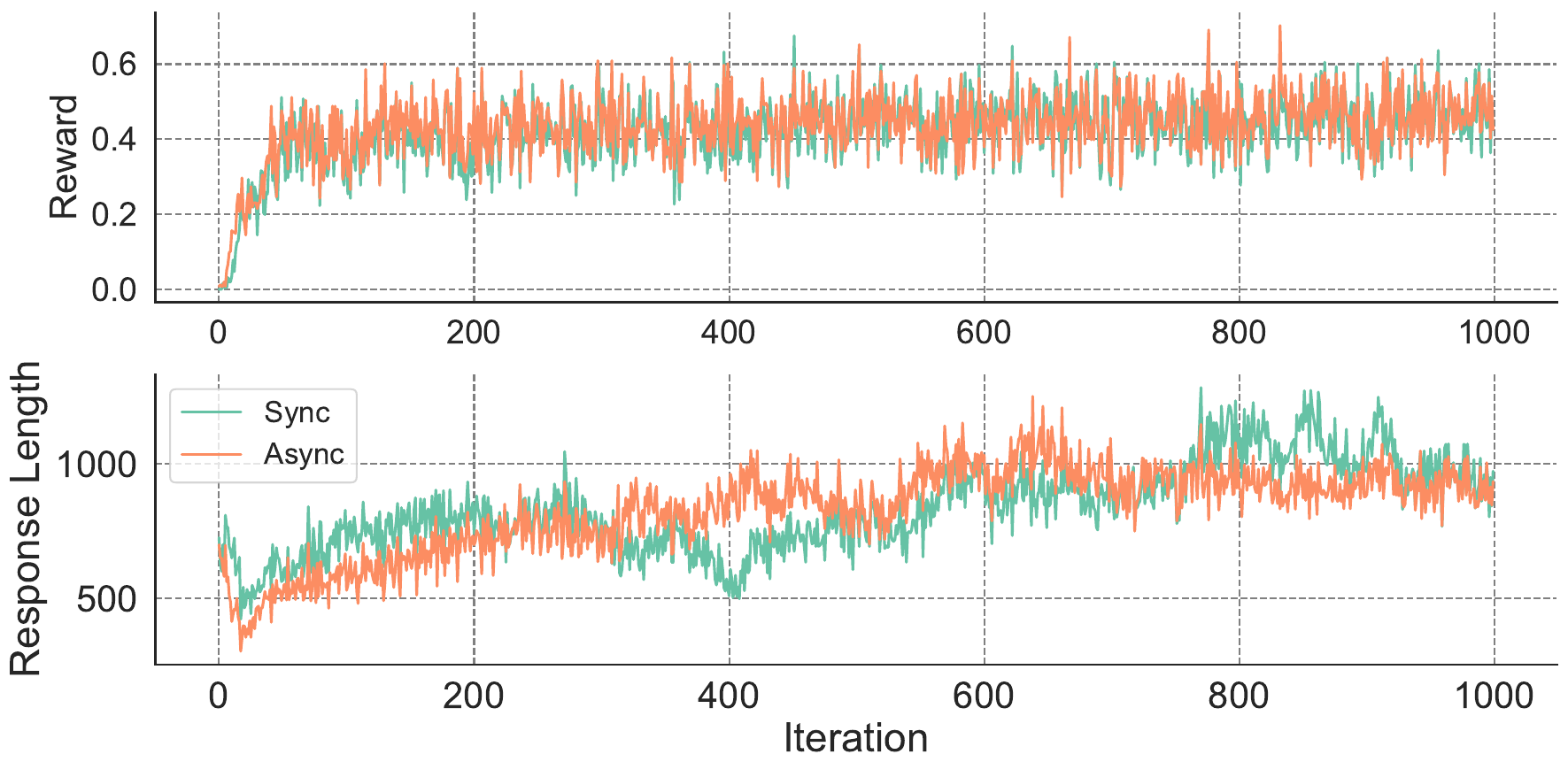}
	\caption{Reward and average response length comparison for the proposed asynchronous RL workflow and vanilla synchronous RL workflow.}
	\label{fig:reward_response_len}
\end{figure}

\subsection{Stability of Asynchronous RL Algorithm}
To evaluate whether the proposed asynchronous RL workflow impacts model performance, we measure the average reward and response length given the same iteration constraint,
as illustrated in Figure \ref{fig:reward_response_len}. The experiments are conducted on a 7B model deployed across 16 NPUs, with the asynchronous workflow alternately enabled and disabled.

Results show negligible differences in reward scores between asynchronous and synchronous workflows, with nearly identical evolutionary trajectories throughout training and convergence to comparable final levels. In contrast, response length variance exhibits slightly different behavioral patterns: after the initial exploratory phase, synchronous mode demonstrates greater fluctuations while the asynchronous approach maintains greater stability. Both methods ultimately converge to stable response lengths, indicating that the asynchronous workflow achieves consistent generative behavior relative to its synchronous baseline.

In conclusion, AsyncFlow preserves the stability of the training and the performance of the model. More importantly, we achieve this while delivering substantially higher throughput, as previously established, presenting a superior balance between efficiency and convergence stability.

%!TEX root = main.tex

\section{Related Work}\label{sec:RelatedWorks}
\subsection{LLM Post-Training Frameworks}
As post-training phases have become indispensable for deploying state-of-the-art large language models to refine their capabilities, numerous frameworks have emerged to efficiently manage and optimize this process. In this study, we classify post-training frameworks into two paradigms: \textbf{task-collocated} and \textbf{task-separated}, based on how reinforcement learning tasks map to physical devices.

Task-collocated frameworks execute both actor rollout and actor update on the same set of devices. Examples include TRL \cite{vonwerra2022trl}, DeepSpeed-Chat \cite{yao2023deepspeed}, NeMo-Aligner \cite{shen2024nemo}, RLHFuse \cite{zhong2025optimizingrlhftraininglarge}, and verl \cite{sheng2024hybridflow}. These frameworks achieve high resource utilization for small-scale post-training.

Task-separated frameworks, in contrast, decouple actor rollout and actor update into distinct hardware resources. Recent high-quality frameworks \cite{seed2025seed,zhong2025streamrl,fu2025areal} demonstrate strong competitiveness in large-scale post-training scenarios. This paradigm aligns with the divergent computational demands of LLM inference (memory-intensive) and training (compute-intensive), enabling better hardware utilization through specialized resource allocation.

\subsection{RLHF Algorithms}
Among RLHF algorithms, PPO \cite{schulman2017proximal} serves as the foundational algorithm, orchestrating four core components: the actor model, the reference model, the reward model, and the critic model. This multi-component architecture introduces intrinsic computational complexity, posing challenges to training efficiency. To mitigate these limitations, recent variants such as GRPO \cite{shao2024deepseekmath} and DAPO \cite{yu2025dapo} simplify the workflow by eliminating specific components. For instance, GRPO removes the critic model by estimating advantages from rewards normalized within a group of responses to the same prompt. These simplifications aim to reduce computational overhead, but may introduce stability issues.

\subsection{LLM Post-Training System Optimization}
Considering the end-to-end time composition of the RL post-training workflow, the actor rollout phase has drawn significant research attention due to its autoregressive inference nature, which fundamentally distinguishes it from training tasks. RLHFuse \cite{zhong2025optimizingrlhftraininglarge} introduces inter-stage fusion strategies that enable concurrent execution of actor rollout with downstream tasks through an automated migration mechanism. StreamRL \cite{zhong2025streamrl} further enhances scheduling flexibility by incorporating a response length predictor. Similarly, the partial rollout technology is introduced in k1.5 \cite{team2025kimi}, which truncates long responses to enable pipelining of downstream tasks without waiting for the full generation. Notably, the rollout phase in post-training lacks strict latency requirement, enabling broader scheduling strategies that maximize the system throughput.

%!TEX root = main.tex
\vspace{-0.17in}
\section{Conclusion}\label{sec:Conclusion}
In this work, we propose a task-separated reinforcement learning framework designed to deliver a modular and scalable post-training capability, especially in large-scale clusters.

To achieve this goal, we first introduce TransferQueue, a streaming data module with panoramic management and distributed storage capabilities. Through dynamically routing the complex data dependencies across training and inference instances, it overcomes bottlenecks imposed by single controller that routing all the training data, where the data dependencies are usually pre-defined. Furthermore, we develop an asynchronous workflow based on a producer-consumer abstraction, which balances training efficiency and convergence stability by allowing one-step parameter asynchronization between \textit{actor rollout} and \textit{actor update}. By overlapping parameter updates with computation tasks, this design minimizes hardware idling across iterations while maintaining algorithmic integrity. For usability design, we implement a set of service-oriented interfaces that provide two-level abstraction of both algorithm workflow and the backend engines, which can effectively bridge the gap between theoretical research and industrial deployment. Extensive experiments demonstrate that AsyncFlow achieves significant throughput improvements over the baseline, exhibiting superior linearity when scaling to larger clusters.

For future work, we identify critical opportunities in rollout system design. Under the constraint that the generated responses sufficiently support downstream tasks, we can stagger the timing of parameter updating for each inference instance to eliminate synchronization barriers during response generation. This approach not only reduces the exposed transition time, but also enables sub-step asynchronization workflow where the staleness threshold between \textit{actor rollout} and \textit{actor update} falls below one training step.

In summary, our study sheds light on the design of task-separated RL frameworks, which demonstrate superior scalability and efficiency in large-scale post-training scenarios.

\section*{Acknowledgments}
We would like to express our sincere gratitude to Jianping Sun, Chao Bai, Long Chen, Benzhe Ning, Xushi Li, Qianqian Cheng, Yayun Ji, Yongwen Li, Wenqin Tan, Yebin Zhang, Peihan Liu, Haoran Dong, Cunle Qian, Xi Chen, Wei Zhou, Yonghao Song, Cangsu Hu, Yinlei Sun, Liangjun Feng, Jiangben Wang and Bo Wang for their invaluable contributions.

\bibliographystyle{plainnat}
\bibliography{references}

\clearpage
\onecolumn
\appendix
\section*{APPENDIX}

\setcounter{figure}{0}
\renewcommand{\thefigure}{A\arabic{figure}}
\begin{figure}[H]
        \centering
	\includegraphics[width=0.48\linewidth]{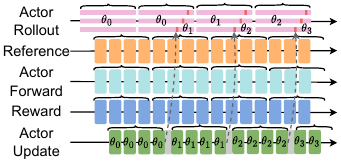}
    \caption{Illustration of sub-step asynchrony workflow by sequentially enabling parameter updates for rollout instances. We leave the implementation of this mechanism for future work.}
    \label{fig:async_algo_future}
\end{figure}

\begin{Code}[H]
\begin{lstlisting}[language=Python]
# In openrlhf/trainer/ray/launcher/PPORayActorGroup, commit id 64c6350a
def async_fit_actor_model(self,):
    # get all instances of actor handlers
    critic_actors = critic_model_group._actor_handlers if critic_model_group else None
    initial_actors = initial_model_group._actor_handlers if initial_model_group else None

    refs = []

    for i, actor in enumerate(self._actor_handlers):
        critic_actor = critic_actors[i % len(critic_actors)] if critic_actors else None
        initial_actor = initial_actors[i % len(initial_actors)] if initial_actors else None

        reward_actors = []
        if not remote_rm_urls:
            for reward_model_group in reward_model_groups:
                actors = reward_model_group._actor_handlers
                reward_actors.append(actors[i % len(actors)])

        # mannually bind the dataflow between actor and critic instances
        # and start training
        refs.append(
            actor.fit.remote(
                critic_model=critic_actor,
                initial_model=initial_actor,
                reward_model=reward_actors,
                remote_rm_url=remote_rm_urls,
                reward_fn=reward_fn,
                vllm_engines=vllm_engines,
                # whether this actor should triger corresponding critic model training
                critic_train_remote=(i < len(critic_actors)) if critic_actor else None,
            )
        )

    return refs
\end{lstlisting}
	\caption{Dataflow example in OpenRLHF (Commit 64c6350a). In this design, we have to manually bind the relationships across DP groups of different RL tasks. Note that this implementation has been replaced by a more efficient alternative in recent releases.}
	\label{code:data_dependency_in_openrlhf}
\end{Code}

\begin{Code}[H]
\begin{lstlisting}[language=Python]
# In class RolloutWorker(BaseWorker):
def generate_sequences(self):
    # define the data columns and initialize the TransferQueue
    experience_consumer_stage = 'actor_rollout'
    experience_columns = ['prompts', 'prompt_length']
    experience_count = self.rl_config.rollout_dispatch_size

    data_loader = self.create_stream_data_loader(
        experience_consumer_stage=experience_consumer_stage,
        experience_columns=experience_columns,
        experience_count=experience_count,
        use_vllm=True,
        pad_to_multiple_of=self.generate_config.infer_tensor_parallel_size,
    )
    data_iter = iter(data_loader)

    for batch_data, index in data_iter:
        # do inference 
        prompts_data = batch_data['prompts']
        responses = self.rollout.generate_sequences(prompts_data)
        # write generated responses to TransferQueue
        self.collect_transfer_queue_data(responses, index)
\end{lstlisting}
	\caption{TransferQueue usage example.}
	\label{code:tq_example}
\end{Code}

\begin{Code}[H]
\begin{lstlisting}[language=Python]
# Base adapter
class RLAdapter:
    pass

class MindSpeedAdapter(RLAdapter):
    def __init__(self, forward_backward_func, model, batches, forward_step):
        self.forward_backward_func = forward_backward_func
        self.forward_step = forward_step
        self.model = model
        self.batches = batches
        ...

    # Abstraction for RL task: compute_log_prob
    def compute_log_prob(self):
        ...
        losses_reduced = self.forward_backward_func(
        forward_step_func=self.forward_step,
        data_iterator=iter(self.batches),
        model=self.model,
        micro_batch_size=self.micro_batch_size,
        forward_only=True,
        collect_non_loss_data=True,
        )
        ...

class VLLMAdapter(RLAdapter):
    pass
\end{lstlisting}
	\caption{Backend-level interface.}
	\label{code:backend_interface_example}
\end{Code}

\end{document}